\documentclass[conference]{IEEEtran}
\usepackage[utf8]{inputenc} 
\usepackage[T1]{fontenc}    
\usepackage{hyperref}       
\usepackage{url}            
\usepackage{booktabs}       
\usepackage{amsfonts}       
\usepackage{nicefrac}       
\usepackage{microtype}      
\usepackage[table]{xcolor}
\usepackage{times}
\usepackage{soul}
\usepackage{url}
\usepackage[utf8]{inputenc}
\usepackage[small]{caption}
\usepackage{url}
\usepackage{graphicx}
\usepackage{wrapfig}
\usepackage{amsmath}
\usepackage{subfigure}
\usepackage{amssymb}
\usepackage{framed}
\usepackage{amsthm}
\usepackage{float}

\usepackage{booktabs}
\usepackage{algorithm}
\usepackage{algorithmic}
\usepackage{setspace}

\newcommand{\numdatasets}{17}

\hypersetup{
    colorlinks=true,
    linkcolor=blue,
    filecolor=black,      
    urlcolor=blue,
    citecolor=blue
}

\ifCLASSINFOpdf
\else
\fi
\begin{document}
%
\title{Compressibility of Distributed \\Document Representations}

\author{\IEEEauthorblockN{Bla\v{z} \v{S}krlj}
\IEEEauthorblockA{Jo\v{z}ef Stefan Institute, Ljubljana, Slovenia \\
and Jo\v{z}ef Stefan International Postgraduate School, Ljubljana, Slovenia \\
Email: blaz.skrlj@ijs.si}
\and
\IEEEauthorblockN{Matej Petkovi\'{c}}
\IEEEauthorblockA{Jo\v{z}ef Stefan Institute, Ljubljana, Slovenia \\
Email: matej.petkovic@ijs.si}}



%


\maketitle

\begin{abstract}
Contemporary natural language processing (NLP) revolves around learning from latent document representations, generated either implicitly by neural language models or explicitly by methods such as doc2vec or similar. One of the key properties of the obtained representations is their dimension. Whilst the commonly adopted dimensions of 256 and 768 offer sufficient performance on many tasks, it is many times unclear whether the default dimension is the most suitable choice for the subsequent downstream learning tasks. Furthermore, representation dimensions are seldom subject to hyperparameter tunning due to computational constraints. The purpose of this paper is to demonstrate that a surprisingly simple and efficient recursive compression procedure can be sufficient to both significantly compress the initial representation, but also potentially improve its performance when considering the task of text classification. Having smaller and less noisy representations is the desired property during deployment, as orders of magnitude smaller models can significantly reduce the computational overload and with it the deployment costs.
We propose \textsc{CoRe}, a straightforward, compression-agnostic framework suitable for representation compression.
The \textsc{CoRe}'s performance is showcased and studied on a collection of 17 real-life corpora from biomedical, news, social media, and literary domains. We explored \textsc{CoRe}'s behavior when considering contextual and non-contextual document representations, different compression levels, and 9 different compression algorithms. Current results based on more than 100{,}000 compression experiments indicate that recursive Singular Value Decomposition offers a very good trade-off between the compression efficiency and performance, making \textsc{CoRe} useful in many existing, representation-dependent NLP pipelines.
\end{abstract}

%
\section{Introduction}
\label{sec:introduction}
Contemporary machine learning methods increasingly rely on the quality of latent representations, produced during e.g., training of deep neural network models or \emph{dimensionality reduction techniques}. The common denominator to many models used in practice throughout science and industry is a rather arbitrary selection of the embedding dimension -- it appears widely accepted that a sufficiently high embedding dimension is preferred (e.g. 256 or 768). However, in many practical scenarios such as the development of embedded systems, mobile and online learning, model compactness is the desired property~\cite{joulin2016fasttextzip,
luo2017thinet}.
There has been research targeted at finding e.g., sufficient neural network architectures for, e.g., mobile deployment \cite{howard2017mobilenets}, and, similarly, the distillation of existing large neural language models \cite{sanh2019distilbert}. Albeit succeeding at their tasks, such research endeavors do not emphasize the actual properties of the obtained representations, but rather the model itself. In recent years, the learned representations have been actively studied, offering insights into how existing representations can be compressed for many practical applications, including image~\cite{damahe2019review} and online text classification~\cite{acharya2019online}. 

Finally, as the manual annotations are potentially expensive, the domain of self-supervised learning explores to what extent can a neural network-based system learn relevant representations without any supervision~\cite{mao2020survey} 
The purpose of this paper is to explore, in the domain of document representation learning, to what extent can an existing representation be efficiently compressed with as little performance loss as possible, in \emph{self-supervised} manner -- without human annotations.

The contributions of this work are:
\textbf{i) We propose \textsc{CoRe}}, an embedding-agnostic methodology for automatic recursive compression of latent document representations.
\textsc{CoRe} can construct up to two orders of magnitude ($768 \rightarrow 8$) smaller representations whilst maintaining the performance within a few percentage points, offering drastic memory consumption reduction for down-stream learning.
\textbf{ii) The proposed methodology is evaluated on~\numdatasets{} real-life data sets}, where we explore to what extent \emph{contextual} and \emph{non-contextual} document representations can be \emph{compressed} with multiple linear and non-linear dimensionality reduction methods, offering novel insights into the compressibility of different types of document representations.
\textbf{iii) We demonstrate that a very efficient SVD-based recursive compression} can offer better-than-initial performance resulting from lower-dimensional representations. 

The remainder of this work is structured as follows. In Section~\ref{sec:related} we discuss the related work.  In Section~\ref{sec:key-idea}, the proposed \textsc{CoRe} methodology is presented, followed by the description of the considered compression algorithms (Section~\ref{sec:algo-compression}), empirical evaluation setting (Section~\ref{sec:experiments}) and the experimental results (Section~\ref{sec:results}). We conclude with discussion and conclusions in Section~\ref{sec:discussion}.
The project's repository is accesible \href{https://github.com/SkBlaz/core}{here}.

\section{Related Work}
\label{sec:related}
The proposed work builds on the ideas from the fields of representation learning and model compression. The notion of representation learning has been considered throughout different sub-domains of machine learning and has received increasing interest in the last years. Neural network-based embedding learning is becoming the prevailing method for obtaining representations (embeddings) of graphs or their nodes~\cite{zhang2018network},  
images~\cite{zerdoumi2018image} and documents~\cite{kesiraju2020learning,reimers-2020-multilingual-sentence-bert}.
In recent years, two main types of document embeddings emerged, namely the ones based on contextual neural language models~\cite{NIPS2017_3f5ee243,Peters:2018} and the non-contextual ones based on earlier word representation learning techniques~\cite{le2014distributed}. 
Commonly used \emph{ad hoc} dimensionalities of the representations of, e.g., 768, 256, and 128 yield satisfying results, however, are commonly not rigorously inspected, possibly due to high computational costs of manual inspection of a model's behavior concerning this hyperparameter. 

Exploration of how language models can be compressed has been an active effort for more than a decade~\cite{talbot-brants-2008-randomized}.
As a result, the development of methods that automatically explore a given representation's properties and attempt to further reduce it is an ongoing research endeavor~\cite{acharya2019online,choi2020universal}. 
With the advent of neural language models, \emph{distillation} re-emerged as a form of model compression~\cite{sun2020contrastive}. Similarly, the pruning of existing models was also shown to be a viable alternative~\cite{liu2018efficient, zhu2017prune}. By selecting the representative subword space, very memory-efficient models can be obtained~\cite{zhao2019extreme}. Similarly, high compressibility of neural language models can be obtained via low-rank approximations~\cite{acharya2019online,chen2018groupreduce}. Recently, attempts to exploit external knowledge to compress transformer-based models such as BERT~\cite{devlin-etal-2019-bert} were also investigated~\cite{sun2019patient}.
Finally, the idea of \emph{self-supervised} learning has recently been explored in the context of language models, albeit originating in the image domain~\cite{liu2020self}. For example, the Albert~\cite{lan2019albert} exploits the inter-sentence coherence to obtain better and more compact language models.
Albeit many compression-related ideas have been proposed and already offered model improvements, to our knowledge no systematic evaluation of different representation compression algorithms has yet been conducted. Furthermore, another rationale for this paper is that it is not clear whether recursive compression is a better option if compared to a direct projection to a lower dimension, which we explored systematically in this work.

\section{Compression of Representations (\textsc{CoRe})}
\label{sec:key-idea}
\begin{figure}[t!]
    \centering
    \includegraphics[width = .85\linewidth]{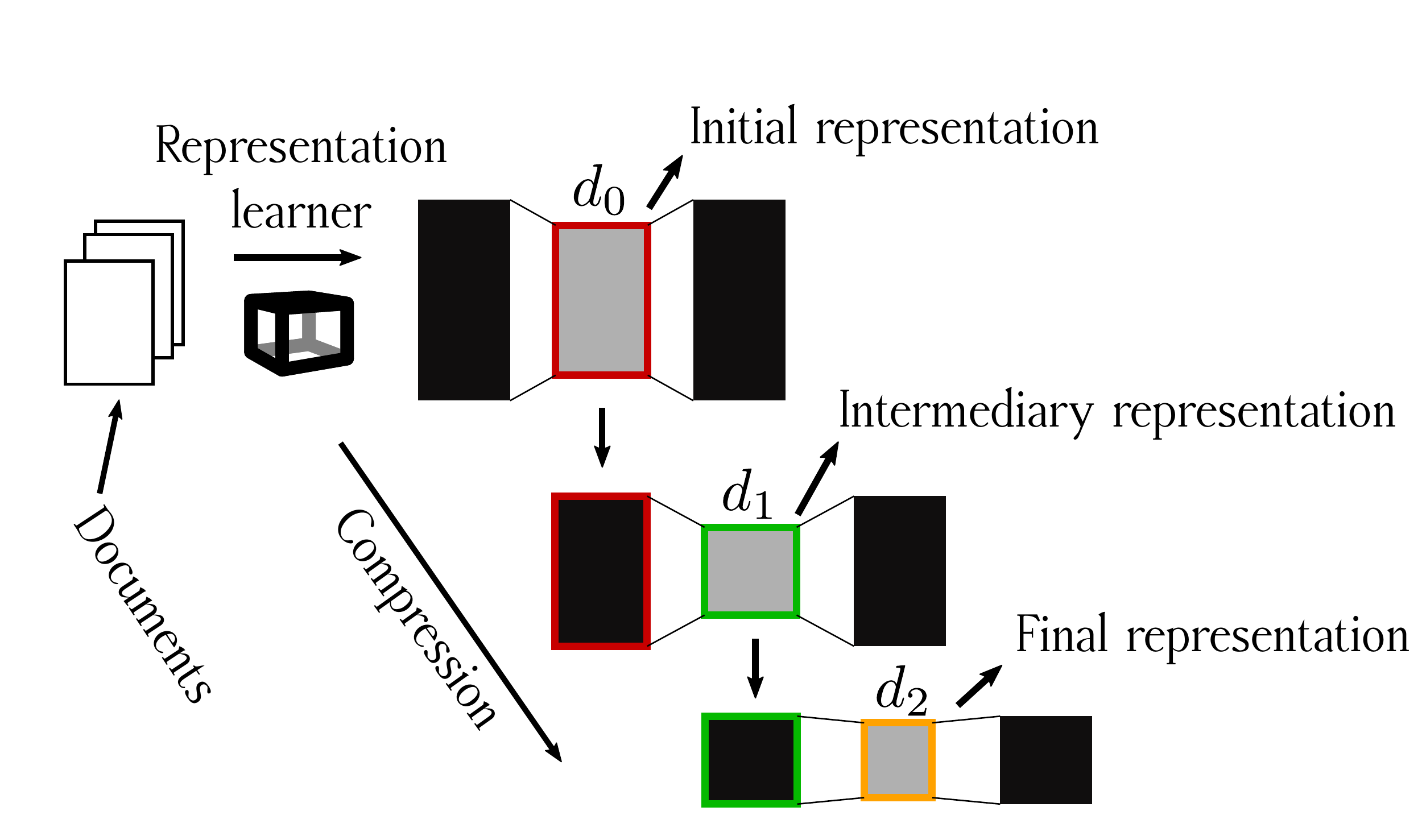}
    \caption{Schematic overview of recursive autoencoding, as explored in this work. The initial representation is recursively compressed ($d_0,d_1,d_2$) to a compact, however expressive representation. This example shows two compression steps only (for readability purposes).}
    \vspace{-0.5cm}
    \label{fig:scheme}
\end{figure}
The proposed approach (schematically shown in Figure~\ref{fig:scheme}) investigates to what extent a given latent representation can be compressed to maintain similar functionality compared to the initially obtained representation. This work builds on the idea of \emph{lossy compression} -- the obtained, more compressed representations can be of lower quality, as long as that quality is sufficient for a given \emph{practical application}. Albeit there exist many compression benchmarks, normally the reconstruction error of the origin e.g., data set is measured. However, as the purpose of \emph{learned} representations is to enable association between input data and e.g., a collection of targets (e.g., genres), the quality of the compressed representations is estimated via extrinsic evaluation with the logistic regression classifier.

\subsection{Recursive autoencoding}
\label{sec:recursive-autoencoding}
Next, we formally describe the incremental reduction of a given representation's dimension -- the key idea of \textsc{CoRe}.  
Let $\boldsymbol{E} \in \mathbb{R}^{|D| \times d_0}$ represent a $d_0$-dimensional representation of a set of documents $D$ -- a corpus. Let $p(d) \in \mathbb{R}$ be the performance of the classifier learned from $d$-dimensional embeddings of the documents. 
The purpose of this paper is to explore how $p(d)$ changes with $d$ and to find 
$d_i$-dimensionional representation, such that $p(d_i)$ 
is at most $\epsilon$ worse (but can be better) than the initial performance $p(d)_0$. 
We further define a function $\textsc{EMB}: \mathbb{R}^{|D| \times d} \rightarrow \mathbb{R}^{|D| \times d_i}$. For example, $\textsc{EMB}(\boldsymbol{E}, 512, 256)$ projects the space $\boldsymbol{E}$ from 512 to 256 dimensions. The key idea of this paper explores the following recurrence relation:
$d_{i + 1} = \max\{d_i / \kappa, \kappa\}$.
The parameter $\kappa$ denotes the dimension reduction factor; the larger the $\kappa$, the more the space is reduced at each step of the optimization. After $i > 0$ steps, \textsc{CoRe} next creates the compressed representation
$\boldsymbol{E}_i = \textsc{EMB}(\boldsymbol{E}_{i - 1},d_{i - 1}, d_i)$, where $\boldsymbol{E}_0 = \boldsymbol{E}$. Here, on each call of the $\textsc{EMB}$ function, we learn how to reduce the dimension of the output of the previous call, starting from the initial data frame $\boldsymbol{E}$ and proceeding towards lower dimensions. 

The key idea 
revolves around \emph{recursive construction} of incrementally smaller representations. Notice that up to this point, we did not discuss the properties of \textsc{EMB} -- the proposed procedure is \emph{agnostic} to the embedding algorithm.

\subsection{Time and Space Complexity}
\label{sec:theoretical}
Let $d_0$ denote the initial representation's dimension. Next, let
$k \sim \log_\kappa (d_0) - 1$
be the number of recursive steps. 
Then, the computational complexity is 
$\mathcal{O} \left( \gamma \cdot |D| \cdot \sum_{i = 1}^k  d_{i - 1} \cdot d_i \right)$, which can be simplified to $\mathcal{O}\left( \gamma \cdot |D| \cdot \kappa^{2k + 1}\right)$, provided that $\kappa$ is lower-bounded by some constant, e.g., $\kappa \geq 2$.
Here, the $d_{i - 1}$ and $d_{i}$ correspond to the dimensions of the input and hidden layer in a single-layer architecture, $\gamma$ to the number of epochs and $|D|$ to the number of documents.
Potentially interesting is also the complexity w.r.t. $\kappa$, i.e. the reduction term. 
By assuming $d_0$ is an exponent of $\kappa$ (commonly the case, e.g., $2^9$), the complexity can also be expressed as $\mathcal{O}(\kappa^{2 k + 1})$, indicating that recursive reduction 
is not much more expensive than its first step, which already requires $\kappa^{2 k + 1}$ steps.
Indeed, the exact value of the above sum 
is $\kappa^3(\kappa^{2k} - 1) / (\kappa^2 - 1)$, thus the proportion of the computations done in the first step of the iteration is approximately $(\kappa^2 - 1)/\kappa^2$ which is always greater than $75\%$ if $\kappa\geq 2$.
 In terms of space complexity, the complexity is bounded as $\mathcal{O}(|D| \cdot d_0)$. The implementation, however, offers an option to hold all intermediary compression steps (and produce them), which results in additional space overhead which was in the considered experiments not problematic.

\section{Representation compression algorithms}  
\label{sec:algo-compression}
In the following section we discuss the representation compression techniques we considered in this work intending to systematically explore the space of low-dimensional document representations. The considered methods are summarised in Table~\ref{tab:methods-all}.
\begin{table}[htb!]
    \centering
    \caption{Overview of the considered compression algorithms. If no reference is given, the algorithm was built for this work.
    }
      \resizebox{.5\textwidth}{!}{
    \begin{tabular}{l|l}
       Compression approach  & Description  \\ \hline
     UMAP \cite{McInnes2018}    & Non-linear compression based on manifold theory \\
     Sparse random projections \cite{li2006very} & Johnson Lindenstrauss lemma-informed projections \\
     Truncated SVD \cite{halko2010finding} & Singular value decomposition \\
     Cluster-aggregation (mean, median, max) & Clustering into $d$ dimensions followed by aggregation \\
     Neural autoencoder - small/large & Neural autoencoders of various complexities\\
     Random subspaces (inspired by \cite{ho1998nearest,Breskvar2018}) & Random subspace of dimension $d$ \\
    \end{tabular}
    }
    \label{tab:methods-all}
\end{table}
The widely adopted methods such as UMAP~\cite{McInnes2018} were shown to perform competitively to most learning-based approaches such as e.g., t-SNE~\cite{tsne}, hence this and similar approaches are not included in this work. As contributions of this work, we implemented the following approaches, of which performance we believe offers additional insights into the representations' properties. The \emph{Neural-small} and \emph{Neural-large} are two differently sized autoencoder architectures. A single layer example can be stated as 
\begin{equation}
   \textsc{AE}(\textbf{D}_r) = \textbf{W}_{\textrm{out}}^T\textrm{SoftSign}(\textrm{BN}(\textrm{Dropout}(\textbf{W}_{\textrm{emb}}^T \textbf{D}_r + b_0))).
    \label{eq:nnet}
\end{equation}
where $\boldsymbol{D}_r$ is a dense representation of $D$ documents.
 The SoftSign activation is defined as $\text{SoftSign}(x) = \frac{x}{1 + |x|}$,
The $\textrm{BN}$ (BatchNorm) is defined as:
$\textrm{BN}(x) = \frac{\textbf{x} - \mathbb{E}[\textbf{x}]}{\sqrt{\mathrm{Var}[\textbf{x}] + \varepsilon}}.$
\noindent The $\varepsilon$ is a small constant required for numeric stability.
The goal of \textsc{AE} is to learn the association
    $\textsc{AE}(\textbf{D}_r) \approx \textbf{D}_r$, $\textsc{AE}: \mathbb{R}^{|D| \times d} \rightarrow \mathbb{R}^{|D| \times d}$.
\noindent To obtain a low-dimensional representation, forward pass is considered only up to the first hidden layer, i.e.,
\begin{equation*}
    \textsc{EMB}(\textbf{D}_r) = \textbf{W}_{\textrm{emb}}^T \textbf{D}_r + b_0 \quad (\textsc{EMB}: \mathbb{R}^{|D| \times d} \rightarrow \mathbb{R}^{|D| \times d_i}).
\end{equation*}
\noindent Note how no activations are employed, ensuring non-activated representations.
The weight updates are considered as follows. The main hypothesis of this work explores whether incremental reconstruction of latent spaces of lower dimension indeed preserves the performance. As such, the autoencoder attempts (at each step) to overfit the representation, and is hence optimized until the loss is $\approx 0$. Note that being able to reconstruct a given input data set with zero error can be related to \emph{lossless compression}; thus, \textsc{CoRe} effectively explores whether incremental steps of theoretically lossless compression yield an useful, low-dimensional (lossy) representation.

Next, we implemented a variant of the random subspaces algorithm, which can be summarised in the following two simple steps. First, randomly select $d_i$ dimensions from the initial representation. Create a subspace-based only on these representations and perform $l_2$ normalization across samples. This approach serves as a very simple/na\"ive baseline.

The third branch of approaches we implemented was aimed at exploring whether the incremental aggregation of detected structures can serve as a simple-to-use compression technique. Here, the two-step procedure operates as follows. First, the KMeans++~\cite{kmeans} algorithm is used on the transposed initial matrix $\textbf{D}_0^T$ to partition the dimensions into $d_i$ clusters. For each detected cluster, an aggregation function is applied. We considered max, mean, and median-based aggregations. Intuitively, this procedure should maintain the key parts of the feature space that are of relevance to maintaining the initial structure. The compression can thus be summarised as:
\begin{equation*}
    \textbf{D}_{i} = \textrm{AGGREGATE-CLUSTERS}(\textrm{KMeans++}(\textbf{D}_{0}^T, d_i)).
\end{equation*}
\noindent This approach explores whether the macro structure of the embeddings offers sufficient expressive power. Note that the recursive version of this algorithm incrementally reduces the feature space (instead of using $\textbf{D}_0$ at each step, it uses the prior representation $\textbf{D}_{i+1}$).
Finally, we included the sparse random projections algorithm as a baseline~\cite{li2006very}.

\begin{table}[htb!]
    \centering
    \caption{Data sets used: their domain and the numbers of documents $D$, tokens $T$, maximal document length $M$, and unique labels $L$.}
    \label{tbl:datasets}
    \resizebox{.45\textwidth}{!}{
    \begin{tabular}{lccccl}
\toprule
                   Data set &  $D$ &  $T$ & $M$ &  $L$ & topic \\
\midrule
                        bbc \cite{greene06icml} &       1406 &   49418 &                  343 &              4 & news topics\\
                   \href{https://www.kaggle.com/deepak711/4-subject-data-text-classification}{subjects} &       1786 &  104972 &                  227 &              4 & text topics\\
                   \href{https://sites.google.com/site/qianmingjie/home/datasets/cnn-and-fox-news}{CNN-news} &       2107 &  144077 &                  605 &              7 & news topics\\
            pan-2017 \cite{rangel2017overview} &       3599 &  575670 &                 1478 &              2 & gender detection\\
                    \href{https://www.kaggle.com/c/detecting-insults-in-social-commentary/data}{insults} &       3946 &   29133 &                  288 &              2& insult detection \\
                  questions \cite{li2002learning}&       5452 &   13278 &                   37 &              6& question categories \\
                    SMSSpam \cite{almeida2011contributions} &       5571 &   17744 &                   88 &              2& SMS messages spam detection\\
             \href{https://appen.com/datasets/medical-sentence-summary-and-relation-extraction/}{MedRelation} &       8361 &   16796 &                   99 &              2&  medical relations detection \\
 AAAI2021 \cite{akhtar2021overview}&       8473 &   41387 &                  149 &              2 & fake news detection \\
                       \href{https://www.kaggle.com/datasnaek/mbti-type}{mbti} &       8675 &  313038 &                  419 &             16& personality type detection (from text) \\
                       \href{https://www.yelp.com/dataset}{yelp} &      10000 &   76482 &                  592 &              5& review classification \\
                 hatespeech \cite{gibert2018hate} &      10557 &   29982 &                  195 &              4& white supremacists hate speech detection\\
                semeval-2019 \cite{zampierietal2019} &      13240 &   39714 &                  155 &              2& offensive speech prediction \\
                   \href{https://www.kaggle.com/guiyihan/text-classification-20}{articles} &      19990 &  285144 &                10692 &             20 & news classification \\
                    sarcasm \cite{misra2019sarcasm} &      28619 &   50241 &                  166 &              2& sarcasm detection \\
                    \href{https://dataworks.iupui.edu/handle/11243/23}{authors} &      53678 &   29997 &                 1008 &             45& Victorian era author detection\\
            drugs-condition \cite{grasser2018aspect}&      70406 &   96879 &                  435 &            150& drug side effects classification \\
\bottomrule
\end{tabular}
}
\end{table}

\section{Empirical evaluation}
\label{sec:experiments}
The experiments were aimed to explore the compressibility of DistilBERT~\cite{sanh2020distilbert} and doc2vec~\cite{le2014distributed} representations concerning both the considered compression level, as well as the compression (autoencoding) algorithm used.

For each compression algorithm (and each data set) we conducted stratified three-fold cross-validation. All experiments were repeated three times, which was possible due to the utilization of the \href{https://www.sling.si/}{SLING} supercomputing infrastructure. Similarly, we report the performance for each compression step. Statistical significance is assessed via critical distance diagrams~\cite{demvsar2006statistical} which compare average ranks achieved with a given method. We conducted the compression experiments on \numdatasets{} different real-life data sets, shown in Table~\ref{tbl:datasets}.


\begin{figure}[htb!]
\centering
    \includegraphics[width=0.98\linewidth]{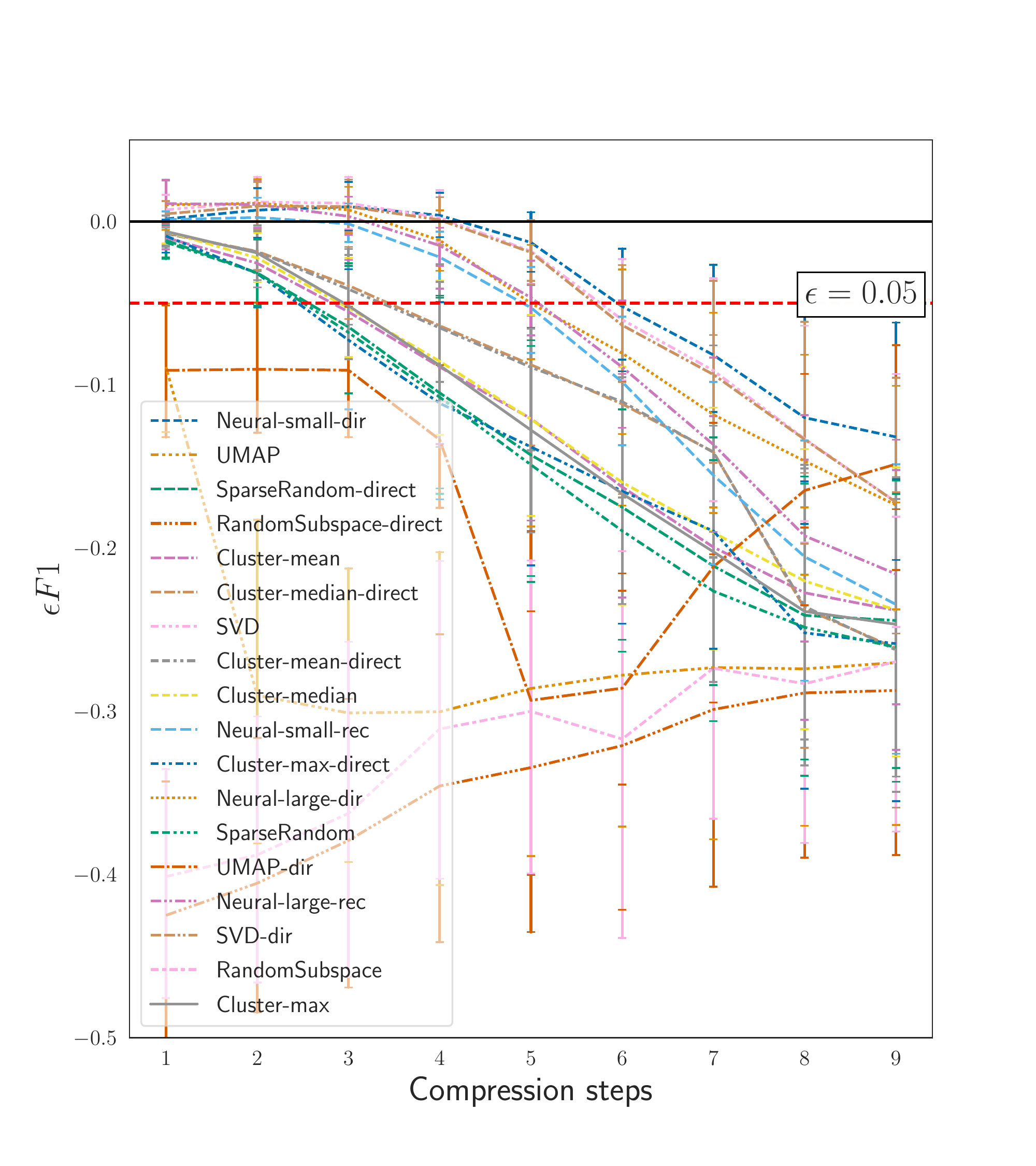}
    \vspace{-0.5cm}
\caption{Overall compression performance for doc2vec-based initial representations and compression (autoencoding) algorithms. The x-axis denotes the number of compression steps (dimension halvings as $\kappa=2$), and y-axis the difference in (micro) F1 performance w.r.t. the initial representation. The method names with ``-dir'' correspond to direct projection into a given dimension (non-recursive compression).}
\vspace{-0.5cm}
\label{fig:summary}
\end{figure}
\subsection{Evaluation of representation quality}
\label{sec:eval}
We are interested in a given compressed representation's relation to the performance of the initial representation. Hence, we introduce $\epsilon F1$, a measure which is computed as:
    $\epsilon F1 = F1_{\textrm{compressed}} - F1_{\textrm{initial}}$.
If $\epsilon F1 > 0$, the compressed representation is \emph{better} by $\epsilon F1$ \textbf{when compared to the original (initial) representation}'s performance. If $\epsilon F1 < 0$, the original representation is better. 
For multi-class problems, we considered the micro $F1$ score. Note that the scores are reported in range $[0, 1]$ and not as percentages. The classifier used in this work is Logistic Regression with regularization parameter C set to 1~\cite{scikit-learn}. The default dimension of representations was set to 768, as commonly adopted in the literature. Each computational job had at most 2 hours to finish on a 16GB (RAM) 8 core virtual machine with no GPU. 

\section{Results}
\label{sec:results}

Due to space limitations, we comment on all representations, but only show the graphs for non-contextual ones (doc2vec).
In Figure~\ref{fig:summary}, the overall impact of compression on the performance can be observed.
 It can be observed that both recursive, as well as non-recursive representations offer sufficient performance when considering multiple compression steps (above the red line of 5\% margin), however, recursive compression offers superior, in the case of doc2vec also better-than-origin representation's performance.
\begin{figure}[htb!]
\centering
    
    \includegraphics[width=0.5\textwidth]{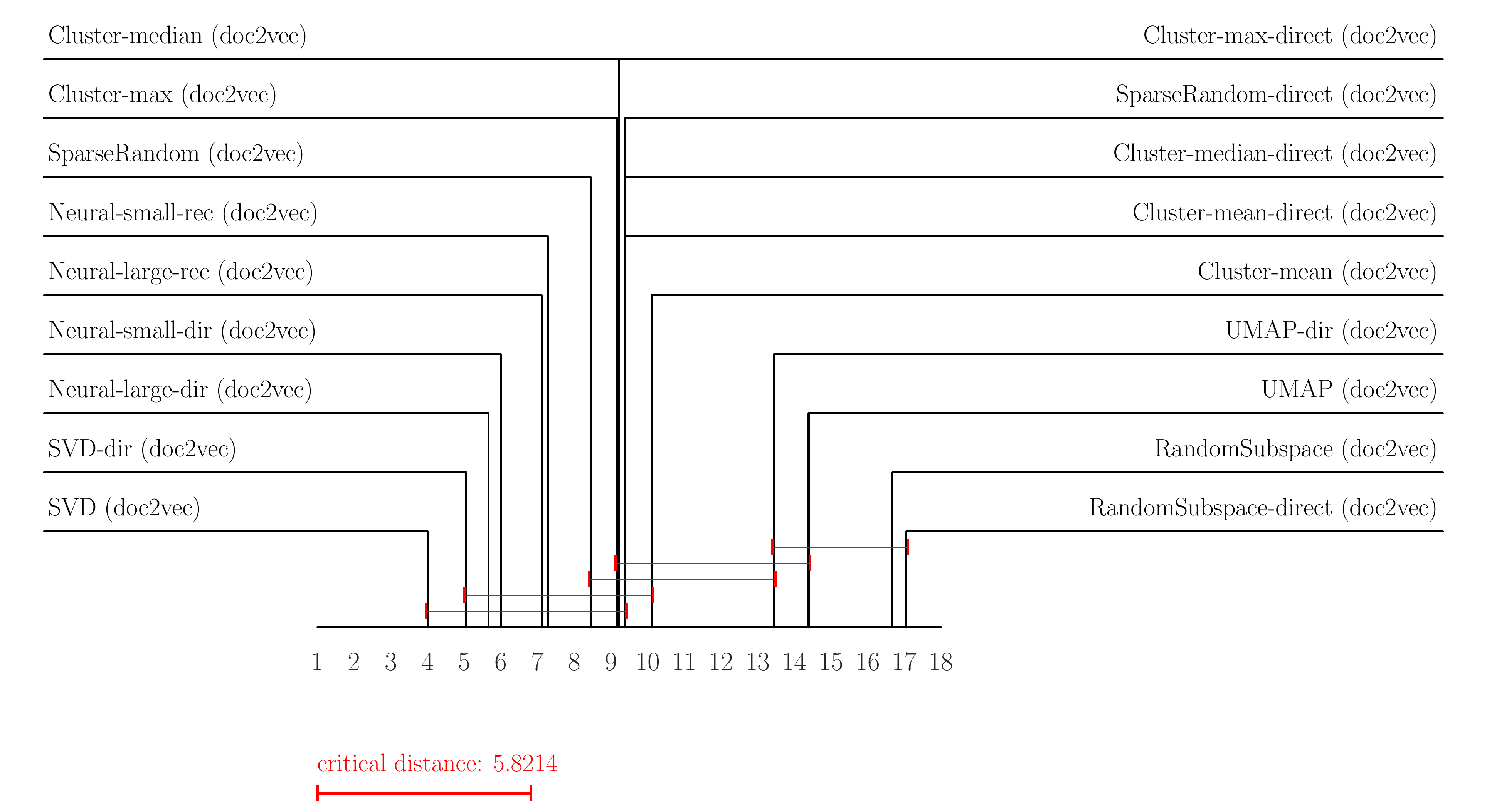}
\caption{doc2vec results. The lines entail statistically (p = 0.05; Friedman-Nemenyi) indistinguishable performance.}
    \label{fig:cd-individual}
\end{figure}
The results indicate that the first few compression steps can even be beneficial for the final representation's quality, however, once very low-dimensional representations are considered, surprisingly, cluster-max-direct yielded the highest-positioned curve, indicating good average performance. This observation only holds for non-contextual representations. For the contextual ones, the SVD-based approaches perform consistently well. 

When comparing all considered approaches (their average performance across the data sets), the first conclusion is that contextual representations (BERT) outperform the non-contextual ones (doc2vec). The other main conclusions are i) that recursive SVD's performance is relatively consistent throughout different levels of compression, ii) that random subspaces are not sufficient for preserving the structure (consistent last places), iii) that the main competitors to SVD are neural autoencoder-based approaches, which are more computationally expensive, albeit being able to perform non-linear decomposition, and iv) that recursive \emph{linear} representation performs well across different compression levels ($\kappa$).

We next present the critical distance diagram~\cite{demvsar2006statistical} in Figure~\ref{fig:cd-individual}. 
The recursive application of SVD consistently offers the best results, regardless of the representation type. Interestingly, the clustering-based reductions were more effective when considering contextual representations (cluster-mean-direct). Overall, however, autoencoder-based representations similarly performed consistently better than e.g., the UMAP-based ones.
The results in tabular form (for SVD-based compression) are given in Table~\ref{tab:svd-comp}.
\begin{table*}[htb!]
    \centering
     \caption{SVD-based compression results across the data sets and $\kappa$ levels. The results where the relative performance w.r.t. the origin representation was $\geq 0$ are highlighted. Note how some of the data sets are easier to compress than others (e.g., \emph{articles} vs. \emph{hatespeech}).}
    \label{tab:svd-comp}
    \resizebox{.98\textwidth}{!}{
    \begin{tabular}{lllllllllllllllllll}
\toprule
                  & Dataset & AAAI2021 &                   CNN-news &            MedRelation &                    SMSSpam &                   articles & authors &                        bbc & drugs &                 hatespeech &                    insults &                       mbti &            pan-2017 &                  questions & sarcasm &                semeval-2019 &                   subjects &                       yelp \\
Compression-representation & Compr. steps &                            &                            &                            &                            &                            &         &                            &                 &                            &                            &                            &                            &                            &         &                            &                            &                            \\
\midrule
SVD (BERT) & 1 &   \cellcolor{green!20}-0.0 &                     -0.004 &                     -0.003 &  \cellcolor{green!20}0.001 &                     -0.018 &  -0.028 &    \cellcolor{green!20}0.0 &          -0.024 &  \cellcolor{green!20}0.013 &  \cellcolor{green!20}0.002 &  \cellcolor{green!20}0.003 &  \cellcolor{green!20}0.004 &                     -0.023 &  -0.019 &  \cellcolor{green!20}0.006 &    \cellcolor{green!20}0.0 &  \cellcolor{green!20}0.022 \\
                  & 2 &                     -0.007 &                     -0.021 &                     -0.036 &  \cellcolor{green!20}0.001 &                      -0.03 &  -0.064 &  \cellcolor{green!20}0.003 &          -0.055 &  \cellcolor{green!20}0.009 &  \cellcolor{green!20}0.013 &  \cellcolor{green!20}0.012 &                     -0.003 &                     -0.051 &  -0.039 &  \cellcolor{green!20}0.013 &    \cellcolor{green!20}0.0 &  \cellcolor{green!20}0.025 \\
                  & 3 &                     -0.017 &                      -0.04 &                      -0.09 &                     -0.001 &                     -0.049 &  -0.111 &    \cellcolor{green!20}0.0 &          -0.108 &  \cellcolor{green!20}0.012 &  \cellcolor{green!20}0.005 &  \cellcolor{green!20}0.025 &  \cellcolor{green!20}0.018 &                     -0.112 &  -0.073 &  \cellcolor{green!20}0.004 &  \cellcolor{green!20}0.002 &   \cellcolor{green!20}0.02 \\
                  & 4 &                     -0.038 &                      -0.08 &                     -0.106 &                     -0.009 &                     -0.093 &  -0.166 &                     -0.009 &          -0.175 &  \cellcolor{green!20}0.011 &                     -0.009 &   \cellcolor{green!20}0.02 &  \cellcolor{green!20}0.011 &                      -0.16 &  -0.108 &                     -0.017 &                     -0.013 &  \cellcolor{green!20}0.019 \\
                  & 5 &                     -0.054 &                     -0.148 &                     -0.134 &                     -0.025 &                     -0.195 &  -0.212 &                      -0.02 &          -0.276 &  \cellcolor{green!20}0.004 &                      -0.01 &  \cellcolor{green!20}0.016 &                     -0.011 &                      -0.21 &   -0.13 &                     -0.021 &                     -0.018 &  \cellcolor{green!20}0.006 \\
                  & 6 &                      -0.07 &                     -0.234 &                     -0.141 &                     -0.026 &                     -0.323 &  -0.242 &                     -0.057 &          -0.358 &                     -0.006 &                      -0.02 &  \cellcolor{green!20}0.021 &  \cellcolor{green!20}0.003 &                     -0.256 &  -0.163 &                      -0.04 &                     -0.027 &                     -0.002 \\
                  & 7 &                     -0.099 &                     -0.324 &                     -0.138 &                      -0.04 &                     -0.458 &  -0.281 &                     -0.119 &          -0.459 &                     -0.008 &                      -0.02 &  \cellcolor{green!20}0.009 &                     -0.052 &                     -0.335 &   -0.24 &                     -0.057 &                     -0.069 &                     -0.023 \\
                  & 8 &                     -0.201 &                     -0.535 &                     -0.158 &                     -0.058 &                     -0.531 &  -0.298 &                     -0.216 &          -0.513 &                     -0.007 &                     -0.031 &  \cellcolor{green!20}0.008 &                     -0.077 &                     -0.425 &   -0.28 &                     -0.073 &                      -0.11 &                     -0.052 \\
                  & 9 &                     -0.207 &                     -0.575 &                     -0.322 &                     -0.073 &                      -0.59 &  -0.298 &                      -0.42 &          -0.516 &                     -0.006 &                     -0.043 &  \cellcolor{green!20}0.006 &                     -0.099 &                     -0.407 &  -0.303 &                     -0.084 &                     -0.195 &                     -0.085 \\ \hline
SVD (doc2vec) & 1 &  \cellcolor{green!20}0.006 &  \cellcolor{green!20}0.014 &  \cellcolor{green!20}0.014 &  \cellcolor{green!20}0.001 &                     -0.012 &  -0.012 &  \cellcolor{green!20}0.001 &          -0.011 &  \cellcolor{green!20}0.002 &  \cellcolor{green!20}0.002 &  \cellcolor{green!20}0.054 &  \cellcolor{green!20}0.016 &                     -0.001 &  -0.001 &  \cellcolor{green!20}0.002 &                     -0.005 &                     -0.001 \\
                  & 2 &  \cellcolor{green!20}0.006 &  \cellcolor{green!20}0.015 &  \cellcolor{green!20}0.014 &  \cellcolor{green!20}0.001 &   \cellcolor{green!20}-0.0 &  -0.019 &    \cellcolor{green!20}0.0 &          -0.039 &  \cellcolor{green!20}0.002 &  \cellcolor{green!20}0.002 &  \cellcolor{green!20}0.106 &  \cellcolor{green!20}0.033 &                     -0.001 &  -0.001 &  \cellcolor{green!20}0.001 &                     -0.005 &                     -0.003 \\
                  & 3 &  \cellcolor{green!20}0.003 &  \cellcolor{green!20}0.036 &  \cellcolor{green!20}0.009 &  \cellcolor{green!20}0.001 &                     -0.005 &   -0.02 &                     -0.006 &          -0.088 &  \cellcolor{green!20}0.002 &  \cellcolor{green!20}0.002 &  \cellcolor{green!20}0.112 &  \cellcolor{green!20}0.043 &                     -0.001 &  -0.006 &                     -0.007 &                     -0.008 &  \cellcolor{green!20}0.004 \\
                  & 4 &                     -0.008 &  \cellcolor{green!20}0.062 &                     -0.003 &  \cellcolor{green!20}0.001 &                     -0.019 &  -0.049 &                     -0.007 &          -0.128 &  \cellcolor{green!20}0.001 &  \cellcolor{green!20}0.001 &  \cellcolor{green!20}0.094 &  \cellcolor{green!20}0.039 &                     -0.001 &  -0.019 &                     -0.028 &                     -0.008 &  \cellcolor{green!20}0.001 \\
                  & 5 &                     -0.014 &  \cellcolor{green!20}0.058 &                     -0.024 &  \cellcolor{green!20}0.001 &                     -0.051 &  -0.166 &                     -0.003 &          -0.162 &  \cellcolor{green!20}0.002 &                     -0.009 &  \cellcolor{green!20}0.039 &   \cellcolor{green!20}0.02 &                     -0.006 &  -0.033 &                     -0.039 &                     -0.005 &                     -0.004 \\
                  & 6 &                     -0.027 &  \cellcolor{green!20}0.054 &                     -0.075 &                     -0.006 &                     -0.149 &  -0.341 &   \cellcolor{green!20}-0.0 &          -0.198 &                     -0.001 &                     -0.013 &                     -0.118 &  \cellcolor{green!20}0.032 &                     -0.022 &  -0.042 &                      -0.04 &                     -0.026 &                     -0.024 \\
                  & 7 &                     -0.032 &  \cellcolor{green!20}0.006 &                     -0.147 &                     -0.015 &                     -0.257 &  -0.518 &                     -0.005 &          -0.256 &  \cellcolor{green!20}0.001 &                      -0.02 &                     -0.174 &                     -0.004 &                     -0.064 &  -0.123 &                     -0.041 &                     -0.036 &                     -0.053 \\
                  & 8 &                     -0.034 &                     -0.182 &                     -0.145 &                     -0.033 &                     -0.428 &  -0.657 &                     -0.066 &          -0.317 &  \cellcolor{green!20}0.004 &                     -0.018 &                      -0.24 &                     -0.185 &                     -0.127 &  -0.164 &                     -0.041 &                     -0.085 &                     -0.117 \\
                  & 9 &                     -0.037 &                     -0.351 &                     -0.149 &                     -0.069 &                     -0.531 &  -0.756 &                     -0.284 &          -0.362 &  \cellcolor{green!20}0.004 &                     -0.021 &                     -0.255 &                      -0.18 &                     -0.208 &  -0.207 &                     -0.041 &                      -0.09 &                     -0.122 \\ \hline
SVD-dir (BERT) & 1 &   \cellcolor{green!20}-0.0 &                     -0.004 &                     -0.003 &  \cellcolor{green!20}0.001 &                     -0.018 &  -0.028 &    \cellcolor{green!20}0.0 &          -0.023 &  \cellcolor{green!20}0.013 &  \cellcolor{green!20}0.002 &  \cellcolor{green!20}0.003 &  \cellcolor{green!20}0.004 &                     -0.022 &  -0.019 &  \cellcolor{green!20}0.006 &    \cellcolor{green!20}0.0 &  \cellcolor{green!20}0.022 \\
                  & 2 &                     -0.008 &                     -0.019 &                     -0.049 &  \cellcolor{green!20}0.001 &                     -0.027 &  -0.064 &    \cellcolor{green!20}0.0 &          -0.056 &  \cellcolor{green!20}0.009 &  \cellcolor{green!20}0.007 &  \cellcolor{green!20}0.013 &                     -0.002 &                     -0.042 &   -0.04 &  \cellcolor{green!20}0.007 &    \cellcolor{green!20}0.0 &  \cellcolor{green!20}0.022 \\
                  & 3 &                     -0.019 &                     -0.042 &                     -0.081 &    \cellcolor{green!20}0.0 &                     -0.048 &   -0.11 &                     -0.003 &          -0.106 &  \cellcolor{green!20}0.009 &  \cellcolor{green!20}0.004 &  \cellcolor{green!20}0.031 &  \cellcolor{green!20}0.018 &                     -0.108 &  -0.074 &  \cellcolor{green!20}0.002 &  \cellcolor{green!20}0.002 &  \cellcolor{green!20}0.016 \\
                  & 4 &                     -0.039 &                     -0.078 &                     -0.105 &                     -0.009 &                     -0.095 &  -0.167 &                     -0.014 &          -0.175 &   \cellcolor{green!20}0.01 &                     -0.007 &  \cellcolor{green!20}0.019 &  \cellcolor{green!20}0.011 &                      -0.16 &  -0.109 &                     -0.016 &                     -0.013 &  \cellcolor{green!20}0.018 \\
                  & 5 &                     -0.056 &                     -0.146 &                     -0.134 &                     -0.027 &                     -0.192 &  -0.212 &                      -0.02 &          -0.276 &  \cellcolor{green!20}0.004 &                     -0.009 &  \cellcolor{green!20}0.017 &                     -0.011 &                     -0.213 &  -0.131 &                     -0.021 &                     -0.018 &  \cellcolor{green!20}0.003 \\
                  & 6 &                     -0.065 &                     -0.233 &                     -0.141 &                     -0.026 &                     -0.323 &  -0.242 &                     -0.058 &          -0.357 &                     -0.005 &                      -0.02 &   \cellcolor{green!20}0.02 &                     -0.001 &                     -0.255 &  -0.162 &                     -0.039 &                     -0.027 &                     -0.004 \\
                  & 7 &                     -0.099 &                     -0.323 &                     -0.138 &                      -0.04 &                     -0.458 &  -0.281 &                     -0.119 &          -0.459 &                     -0.008 &                      -0.02 &  \cellcolor{green!20}0.009 &                     -0.054 &                     -0.333 &   -0.24 &                     -0.057 &                     -0.069 &                     -0.024 \\
                  & 8 &                     -0.201 &                     -0.535 &                     -0.158 &                     -0.058 &                     -0.531 &  -0.299 &                     -0.216 &          -0.513 &                     -0.007 &                     -0.031 &  \cellcolor{green!20}0.008 &                     -0.078 &                     -0.424 &  -0.279 &                     -0.073 &                      -0.11 &                     -0.053 \\
                  & 9 &                     -0.207 &                     -0.575 &                     -0.322 &                     -0.073 &                      -0.59 &  -0.298 &                      -0.42 &          -0.516 &                     -0.006 &                     -0.043 &  \cellcolor{green!20}0.006 &                     -0.099 &                     -0.407 &  -0.303 &                     -0.084 &                     -0.195 &                     -0.086 \\ \hline
SVD-dir (doc2vec) & 1 &  \cellcolor{green!20}0.007 &                     -0.012 &                     -0.008 &  \cellcolor{green!20}0.003 &                     -0.005 &  -0.016 &  \cellcolor{green!20}0.004 &          -0.009 &  \cellcolor{green!20}0.001 &    \cellcolor{green!20}0.0 &  \cellcolor{green!20}0.069 &   \cellcolor{green!20}0.03 &  \cellcolor{green!20}0.001 &  -0.003 &                     -0.003 &    \cellcolor{green!20}0.0 &  \cellcolor{green!20}0.002 \\
                  & 2 &  \cellcolor{green!20}0.006 &                     -0.011 &                     -0.008 &  \cellcolor{green!20}0.003 &  \cellcolor{green!20}0.007 &  -0.023 &  \cellcolor{green!20}0.003 &          -0.036 &  \cellcolor{green!20}0.001 &    \cellcolor{green!20}0.0 &  \cellcolor{green!20}0.125 &  \cellcolor{green!20}0.052 &    \cellcolor{green!20}0.0 &  -0.004 &                     -0.003 &    \cellcolor{green!20}0.0 &  \cellcolor{green!20}0.002 \\
                  & 3 &  \cellcolor{green!20}0.004 &  \cellcolor{green!20}0.008 &                     -0.014 &  \cellcolor{green!20}0.003 &  \cellcolor{green!20}0.004 &  -0.025 &                     -0.004 &          -0.086 &    \cellcolor{green!20}0.0 &    \cellcolor{green!20}0.0 &  \cellcolor{green!20}0.132 &  \cellcolor{green!20}0.059 &  \cellcolor{green!20}0.001 &  -0.008 &                     -0.011 &                     -0.002 &  \cellcolor{green!20}0.007 \\
                  & 4 &                     -0.007 &  \cellcolor{green!20}0.037 &                     -0.022 &  \cellcolor{green!20}0.003 &                     -0.011 &  -0.054 &   \cellcolor{green!20}-0.0 &          -0.126 &                     -0.001 &                     -0.002 &  \cellcolor{green!20}0.112 &  \cellcolor{green!20}0.056 &    \cellcolor{green!20}0.0 &   -0.02 &                     -0.032 &                     -0.003 &  \cellcolor{green!20}0.003 \\
                  & 5 &                     -0.016 &  \cellcolor{green!20}0.033 &                     -0.048 &  \cellcolor{green!20}0.003 &                     -0.041 &  -0.169 &  \cellcolor{green!20}0.004 &          -0.158 &  \cellcolor{green!20}0.001 &                     -0.009 &  \cellcolor{green!20}0.059 &  \cellcolor{green!20}0.038 &                     -0.003 &  -0.034 &                     -0.043 &  \cellcolor{green!20}0.004 &                     -0.004 \\
                  & 6 &                     -0.028 &   \cellcolor{green!20}0.03 &                     -0.098 &                     -0.003 &                     -0.143 &  -0.345 &                     -0.003 &          -0.201 &                     -0.002 &                     -0.015 &                     -0.101 &   \cellcolor{green!20}0.05 &                     -0.021 &  -0.045 &                     -0.045 &                     -0.022 &                     -0.022 \\
                  & 7 &                     -0.033 &                     -0.013 &                     -0.158 &                     -0.012 &                     -0.248 &  -0.522 &                     -0.004 &          -0.256 &  \cellcolor{green!20}0.001 &                      -0.02 &                     -0.157 &  \cellcolor{green!20}0.013 &                      -0.06 &   -0.12 &                     -0.044 &                     -0.033 &                      -0.05 \\
                  & 8 &                     -0.033 &                     -0.206 &                     -0.162 &                     -0.031 &                     -0.419 &  -0.661 &                     -0.062 &          -0.315 &  \cellcolor{green!20}0.004 &                     -0.019 &                     -0.219 &                     -0.166 &                     -0.124 &  -0.166 &                     -0.044 &                     -0.076 &                     -0.111 \\
                  & 9 &                     -0.038 &                     -0.372 &                     -0.167 &                     -0.065 &                     -0.524 &   -0.76 &                     -0.279 &           -0.36 &  \cellcolor{green!20}0.004 &                     -0.022 &                     -0.236 &                     -0.163 &                     -0.208 &  -0.211 &                     -0.044 &                     -0.079 &                     -0.115 \\
\bottomrule
\end{tabular}

    }
\end{table*}
Tabular results are in alignment with Figure \ref{fig:cd-individual}. 
We also observed both drastic performance reductions relatively quickly (e.g., the \emph{authors} data set), but also very gradual decline (\emph{hatespeech}). Interestingly, the \emph{drugs-condition} data set, comprised of very long documents, was very hard to compress, resulting always in $\epsilon F1 < 0$.  

\begin{figure}[htb!]
    \centering
    \includegraphics[width = .89\linewidth]{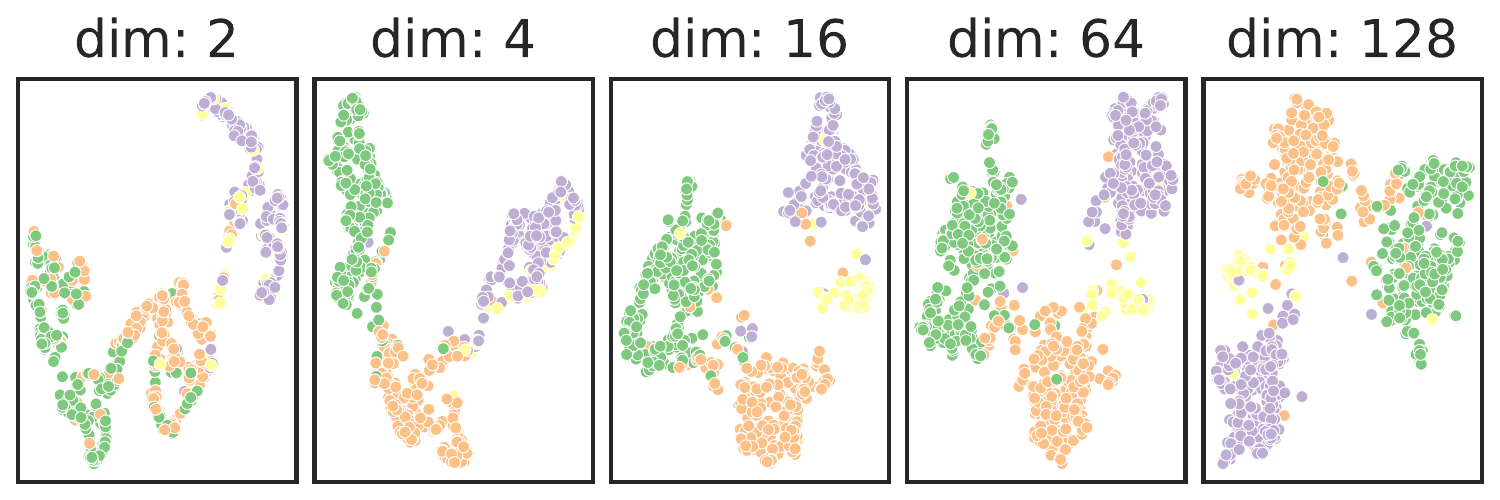}
    \caption{Incremental reduction of dimension with recursive SVD and cluster visualization in 2D (\textit{bbc}). 
    The 2D visualizations were for higher dimensions obtained with default UMAP parameters.}
    \vspace{-0.2cm}
    \label{fig:ablation-bbc}
\end{figure}

As an ablation study, we visualized the
\emph{bbc} data set projected to 2D from different higher dimensions (compression levels) in Figure~\ref{fig:ablation-bbc}.
The observation indicating that in some data sets very low dimensions preserve the initial class structure if compressed in recursive manner with \textsc{CoRe} indicate, that many real-life problems are potentially over-represented in terms of the dimension commonly considered, and could be analysed via much more compact (and efficient) representations.


\section{Discussion and Conclusions}
\label{sec:discussion}
In Section~\ref{sec:results}, we demonstrated that both contextual and non-contextual representations can be compressed by a factor of up to 100x (e.g., on the \emph{bbc} data set), whilst 
maintaining the representation's classification performance. The observed compressibility varied mostly with respect to the number of documents considered for learning. In this work, we also observed the net positive $\epsilon F1$ (Table~\ref{tab:svd-comp}).
In terms of the compression performance, we observed that linear recursive compression performed surprisingly well (\textsc{SVD}). Overall, the non-contextual representations (doc2vec-based) could be more easily compressed -- in some cases (Figure~\ref{fig:summary}), only tens of dimensions were needed to retain a net positive $\epsilon F1$. 
On the contrary, contextual representations are according to our experiments harder to compress. 
Overall, however, tens of dimensions were identified by some of the best-performing methods as enough to retain $5\%\; \epsilon F1$ margin, which could already be of practical relevance. An example where lower-dimensional representations could be practical and speed up the overall learning process are AutoML systems.

As the purpose of this work was to explore whether a technique requiring minimal or zero hyperparameter tunning can already improve the performance, we believe that according to the current results, recursive SVD-based compression is the most suitable one, 
even though 
UMAP and neural methods have the potential to perform even better (at the cost of the additional hyperparameter tuning). 

Further work includes investigation of a larger collection of neural language model-based representation to confirm the current results. 
Finally, we identified as an open problem the \emph{margin identification}, i.e., how to efficiently determine $\kappa$ without computing all intermediary compressions. 
Identifying $\kappa$ upfront could offer automation of the compression, and, when combined with recursive SVD which required zero hyperparameter tunning, provide a powerful \emph{post-hoc} method for optimizing representation-based learners.

\section*{Acknowledgment}
\noindent We thank Tom\'{a}\v{s} Mikolov for his invaluable comments.
The work was supported by the Slovenian Research Agency (young researcher grant (B\v{S}), grant P2-0103), and by the European Commision 
(grants 825153, 
N2-0078, 
952215, 
825619). 



\bibliographystyle{IEEEtran}
\bibliography{references.bib}
%

\end{document}